\documentclass[10pt,twocolumn,letterpaper]{article}

\usepackage{times}
\usepackage{epsfig}
\usepackage{graphicx}
\usepackage{amsmath}
\usepackage{amssymb}
\usepackage{color}
\usepackage{booktabs}   
\usepackage{multirow}
\usepackage{subfig}
\usepackage{mathabx}

\usepackage[pagebackref=true,breaklinks=true,letterpaper=true,colorlinks,bookmarks=false]{hyperref}

\newcommand{\commented}[1]{}
\begin{document}

\title{S3Pool: Pooling with Stochastic Spatial Sampling}

\author{Shuangfei Zhai\\
Binghamton University\\
{\tt\small szhai2@binghamton.edu}
\and
Hui Wu\\
IBM T.J. Watson Research Center\\
{\tt\small wuhu@us.ibm.com}
\and 
Abhishek  Kumar\\
IBM T.J. Watson Research Center\\
{\tt\small abhishk@us.ibm.com}
\and 
Yu  Cheng\\
IBM T.J. Watson Research Center\\
{\tt\small chengyu@us.ibm.com}
\and 
Yongxi Lu\\
University of California, San Diego\\
{\tt\small yol070@ucsd.edu}
\and
Zhongfei (Mark) Zhang\\
Binghamton University\\
{\tt\small zhongfei@cs.binghamton.edu}
\and 
Rogerio  Feris\\
IBM T.J. Watson Research Center\\
{\tt\small rsferis@us.ibm.com}
}
\date{}

\maketitle

\begin{abstract}
Feature pooling layers (e.g., max pooling) in convolutional neural networks (CNNs) serve the dual purpose
of providing increasingly abstract representations as well as yielding computational savings
in subsequent convolutional layers.
We view the pooling operation in CNNs as a two-step procedure: first, a pooling window (e.g., $2\times 2$) slides over the feature map with stride one which leaves the spatial resolution intact, 
and second, downsampling is performed by selecting one pixel from each non-overlapping pooling
window in an often uniform and deterministic (e.g., top-left) manner.
Our starting point in this work is the observation that this regularly spaced downsampling arising from non-overlapping windows,
although intuitive from a signal processing perspective (which has the goal of signal reconstruction),
is not necessarily optimal for \emph{learning} (where the goal is to generalize).                    
We study this aspect and propose a novel pooling strategy with stochastic spatial sampling (S3Pool),
where the regular downsampling is replaced by a more general stochastic version.     
We observe that this general stochasticity acts as a strong regularizer, and can also be seen
as doing implicit data augmentation by introducing distortions in the feature maps. 
We further introduce a mechanism to control the amount of distortion to suit different datasets and architectures.
To demonstrate the effectiveness of the proposed approach,
we perform extensive experiments on several popular image classification benchmarks, 
observing excellent improvements over baseline models \footnote{Experimental code is available at https://github.com/Shuangfei/s3pool}.
\end{abstract}

\begin{figure*}[tbh]
\begin{minipage}{0.5\textwidth}
\centering
\includegraphics[scale=0.5]{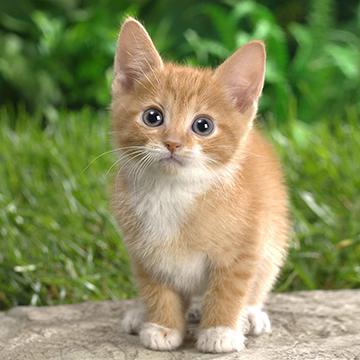}
\end{minipage}
\begin{minipage}{0.5\textwidth}
\centering
\includegraphics[scale=0.5]{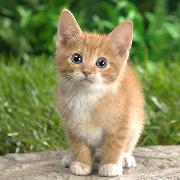}
\includegraphics[scale=0.5]{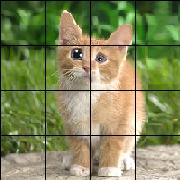}
\includegraphics[scale=0.5]{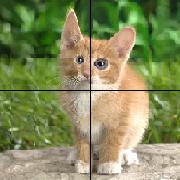}
\includegraphics[scale=0.5]{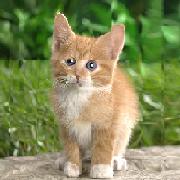}
\end{minipage}
\caption{Illustration of the effect of different downsampling strategies. Left panel: the image before downsampling. Right panel from top left to bottom right: uniform downsampling, stochastic spatial downsampling with the grid size equivalent to a quarter of the image width/height, half of the image width/height, and the image width/height, respectively.}
\label{fig:cat}
\end{figure*}

\section{Introduction}
The use of pooling layers (max pooling, in particular) in deep convolutional neural networks (CNNs) is critical for their success in modern object recognition systems. In most of the common implementations, each pooling layer downsamples the spatial dimensions of feature maps by a factor of $s$ (e.g., 2). This not only reduces the amount of computation required by the time consuming convolution operation in subsequent layers of the network, it also facilitates the higher layers to learn more abstract representations by looking at larger receptive fields. 

In this paper, we provide new insights into the design of the pooling operation by viewing it as a two-step procedure. In the first step, a pooling window slides over the feature map with stride size $1$ producing the pooled output; in the second step, spatial downsampling is performed by extracting the top-left corner element of each disjoint $s\times s$ window, resulting in a feature map with $s$ times smaller spatial dimensions. Our starting point in this work is the observation that although this uniformly spaced spatial downsampling is reasonable from a signal processing perspective which aims for signal reconstruction 
\cite{shannon1949communication}
 and is also computationally friendly, it is not necessarily the optimal design for the purpose of \emph{learning} which aims for generalization to unseen examples\footnote{Uniform sampling has also been examined in the Signal Processing literature, e.g., J. R. Higgins writes \cite{john1996sampling}: \emph{``What is special about equidistantly spaced sample points?''; and then finding that the answer is ``Within certain limitations, nothing at all.''}}. 

\commented{The deterministic nature of downsampling looses valuable information. For example, for a pooling layer of stride $2\times 2$, 3 equally plausible uniformly downsampled feature maps are discarded. For a CNN with $L$ pooling layers, this looses $2^{2L} - 1$ possible combinations of feature maps shown to the model. From a machine learning perspective, uniform spatial downsampling reduces the number of training examples drastically, thus increases the chance of overfitting.} 

\begin{figure}[!t]
\centering
\subfloat[Max pooling, pooling window $k=2$, stride $s=2$]{\includegraphics[width=0.9\columnwidth]{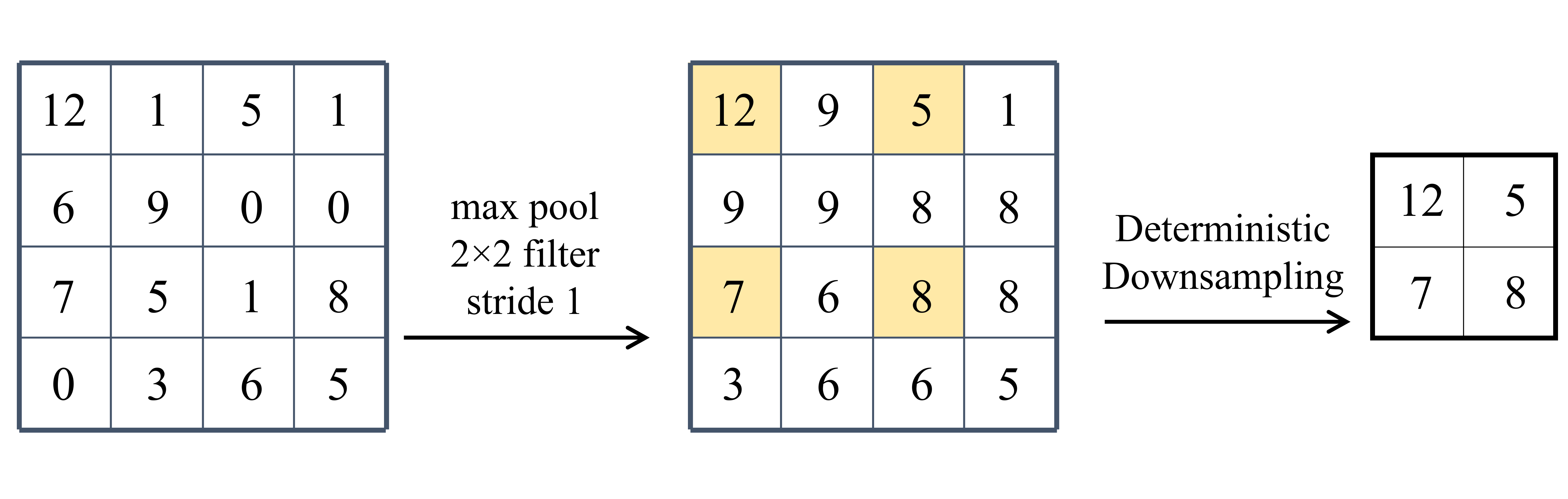}}\\
\subfloat[Stochastic pooling \cite{zeiler2013stochastic}, pooling window $k=2$, stride $s=2$ ]{\includegraphics[width=0.9\columnwidth]{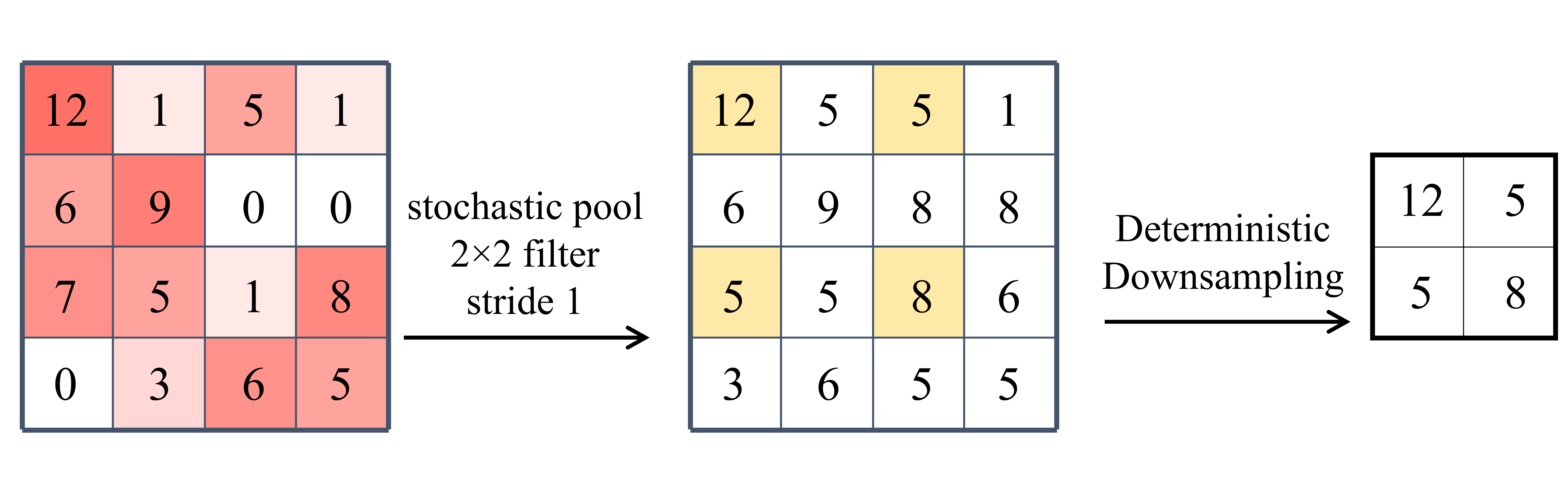}}\\
\subfloat[S3Pool, pooling window $k=2$, stride $s=2$, grid size $g=2$]{\includegraphics[width=0.9\columnwidth]{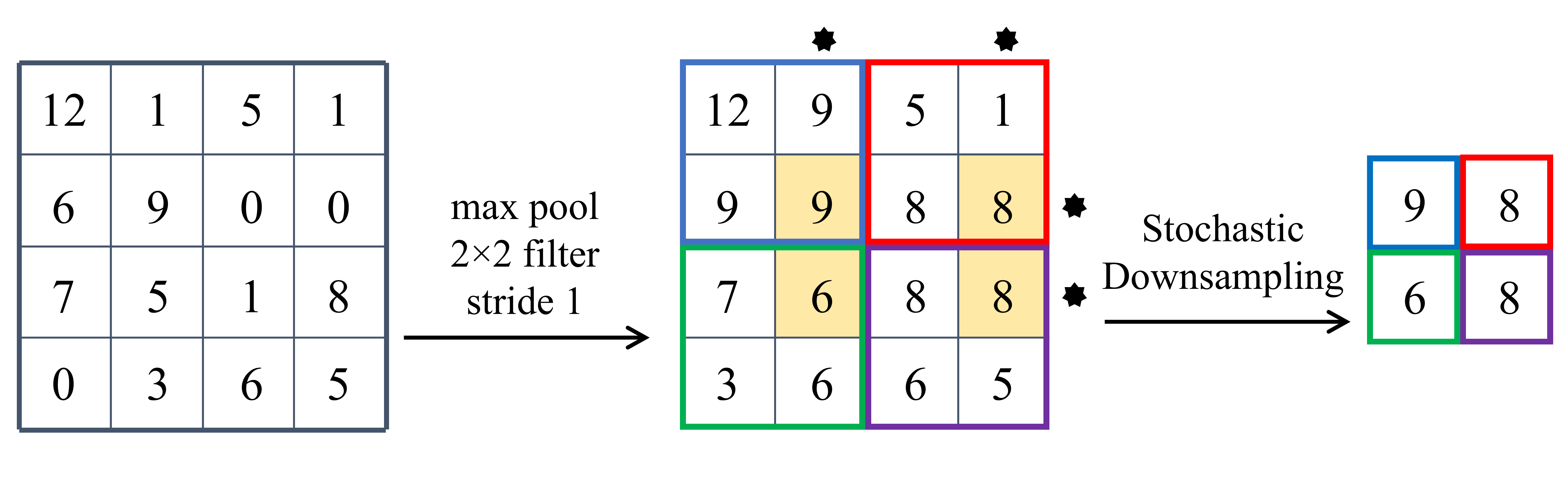}}
\caption{Comparison of different pooling methods (best seen in color). Max pooling (a) consists of two steps, selecting the
activation inside each pooling region and spatial downsampling, where both steps are deterministic. Stochastic pooling~\cite{zeiler2013stochastic}
adapts the first step by choosing the activation with a stochastic procedure (b). While our method modifies
the second step by randomly selecting rows and columns from each spatial grid (c).
}
\label{fig:diagram1}
\end{figure}

Motivated by this observation, we introduce and study a novel pooling scheme, named \emph{S3Pool}, where the second step (downsampling) is modified to a stochastic version. For a feature map with spatial dimensions $h\times w$, S3Pool begins with partitioning it into $p$ vertical and $q$ horizontal strips, with $p=\frac{h}{g}$, $q=\frac{w}{g}$ and $g$ being a hyperparameter named grid size. It then randomly selects $\frac{g}{s}$ rows and $\frac{g}{s}$ columns for each horizontal and vertical strip, respectively, to obtain the final downsampled feature map of size $\frac{h}{s}\times \frac{w}{s}$.
\commented{To put it simple, for a pooling layer with stride $s\times s$ and feature map size $h\times w$, S3Pool randomly selects $\frac{h}{s}$ rows and $\frac{w}{s}$ columns as the downsampled feature map.} 
Compared to the downsampling used in standard pooling layers, S3Pool performs a spatial downsampling that is  \emph{stochastic} and hence is highly likely to be \emph{non-uniform}.
\commented{Figure \ref{fig:cat} visualizes the effect of stochastic spatial downsampling on the input image,
whereas Figure \ref{fig:diagram1} illustrates its differences from standard max pooling.}
The stochastic nature of S3Pool enables it to produce different feature maps at each pass for the same training examples, which amounts to implicitly performing a sort of data augmentation \cite{simard2003best}, but at intermediate layers. Moreover, the non-uniform characteristics of S3Pool further extends the space of possible downsampled feature maps, which produces spatially distorted downsampled versions at each pass. The grid size $g$ provides a handle for controlling the amount of distortion that S3Pool introduces, which can be used to adapt to CNNs with different designs, and different datasets. Overall, S3Pool acts as a strong regularizer by performing `virtual' data augmentation at each pooling layer, and greatly enhances a model's generalization ability as observed in our empirical study.

Practically, S3Pool does not introduce any additional parameters, and can be plugged in place of any existing pooling layers. We have also empirically verified that S3Pool only introduces marginal computational overheads during training time (evaluated by time per epoch). During test time, S3Pool can either be reduced to standard max pooling, or be combined with an additional average pooling layer for a slightly better approximation of the stochastic downsampling step. In our experiments, we show that S3Pool yields excellent results on three standard image classification benchmarks, with two state-of-the-art architectures, namely network in network \cite{nin}, and residual networks \cite{residualnet}. We also extensively experiment with different data augmentation strategies, and show that under each setting, S3Pool is able to outperform other counterparts such as dropout \cite{dropout} and stochastic pooling \cite{zeiler2013stochastic}. 

\section{Related Work}

The idea of spatial feature pooling dates back to the seminal work by Hubel and Wiesel \cite{Hubel62} about complex cells in the mammalian visual cortex and the early CNN architectures developed by Yann Lecun \textit{et al.} \cite{Lecun98}. Prior to the re-emergence of deep neural networks in computer vision, different approaches based on bag-of-words and fisher vector coding also had spatial pooling as an essential component of the visual recognition pipeline, e.g., through orderless bag-of-features \cite{Csurka04,Kristen05}, spatial pyramid aggregation \cite{Lana06}, or task-driven feature pooling \cite{TDP15}.

In modern CNN architectures, spatial pooling plays a fundamental role in achieving invariance (to some extent) to image transformations, and produces more compact representations for efficient processing in subsequent layers. Most existing methods rely on {\em max} or {\em average} pooling layers. Hybrid pooling \cite{Lee16,Patch15} combines different types of pooling into the same network architecture. Stochastic pooling \cite{zeiler2013stochastic} randomly picks the activation within each pooling region according to a multinomial distribution. Max-out networks \cite{maxout,pmaxout} perform pooling across different feature maps. Spatial pyramid pooling \cite{he2014spatial} aggregates features at multiple scales, and is usually applied to extract fixed-length feature vectors from region proposals for object detection. Fractional pooling \cite{fractional} proposes to use pooling strides of less than 2 by applying mixed pooling strides of 1 and 2 at different locations. Learning-based methods for spatial feature pooling have also been proposed \cite{BengioPool14,Coates11}.

As discussed previously, we view pooling as two distinct steps and propose stochastic spatial sampling as a novel solution that has not been investigated in previous work, to the best of our knowledge. Our approach is simple to implement, very efficient, and complementary to most of the techniques discussed above.

\begin{figure}[!tbh]
\centering
\subfloat[stride $s=2$, grid size $g=4$]{\includegraphics[scale=0.3]{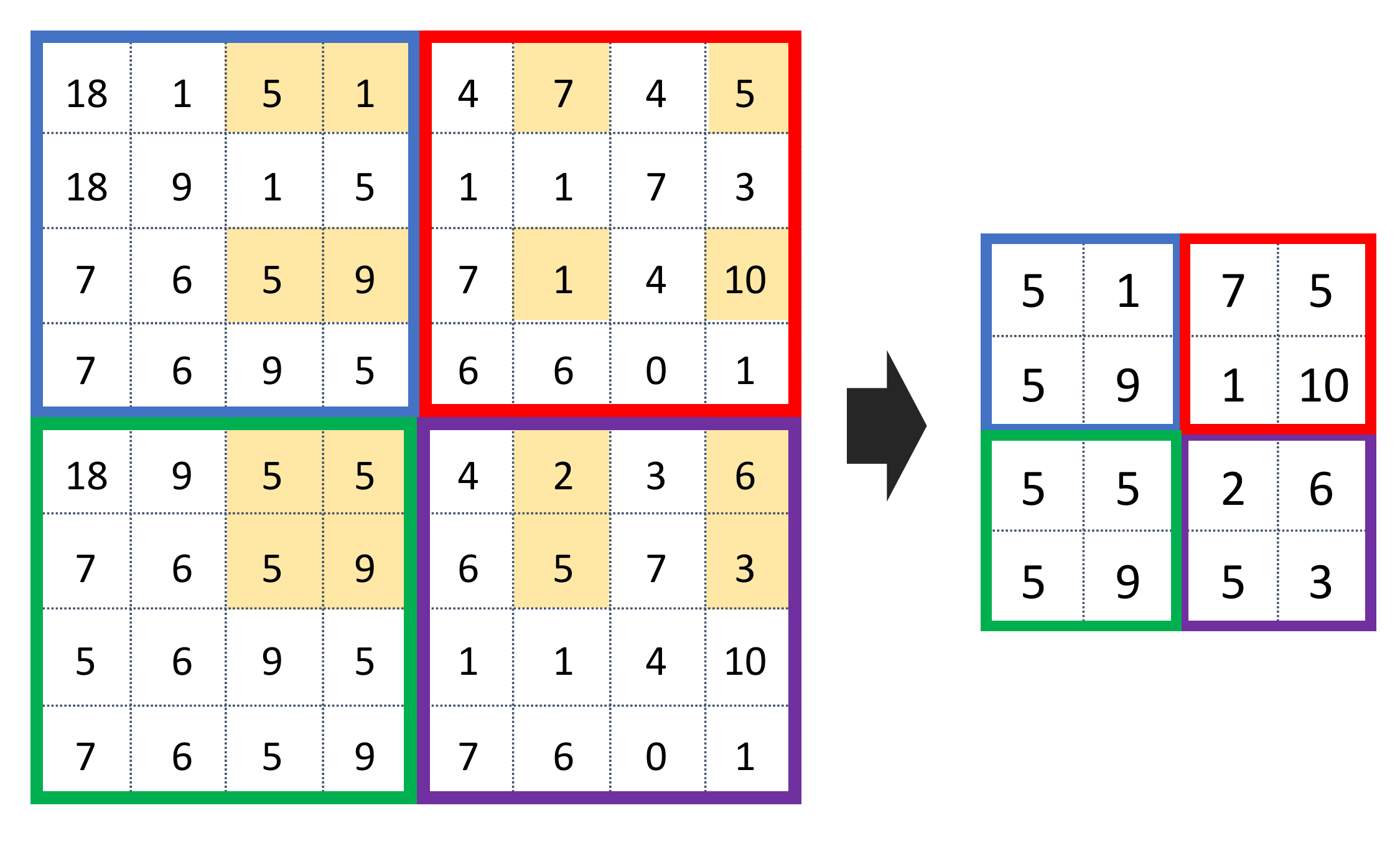}}\\
\subfloat[stride $s=2$, grid size $g=2$]{\includegraphics[scale=0.3]{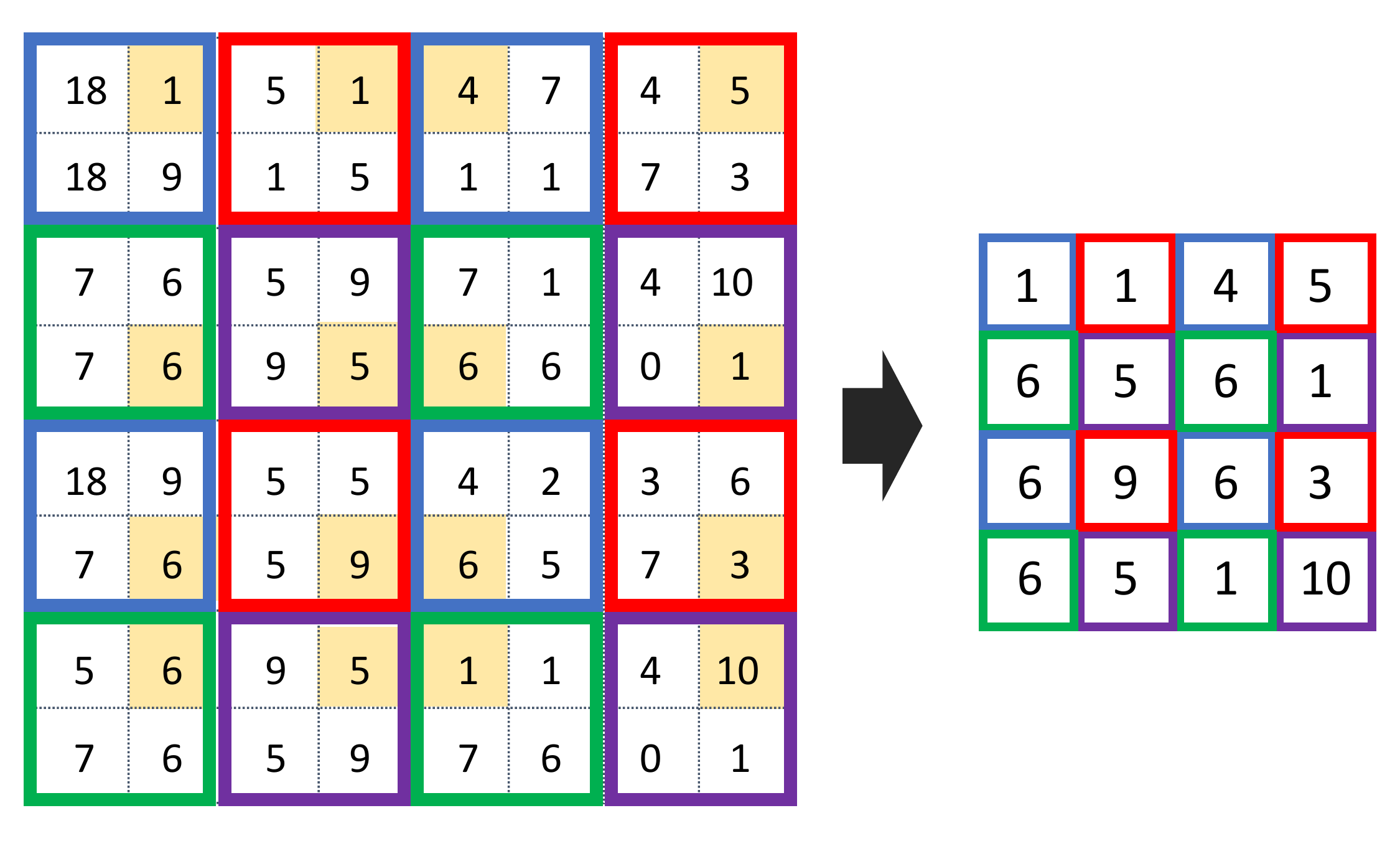}}
\caption{Controlling the amount of distortion/stochasticity by changing the grid size $g$ in the stochastic downsampling step (best seen in color).
}
\label{fig:diagram2}
\end{figure}

\section{Model Description}
\subsection{A Two-Step View of Max Pooling}
Max pooling is perhaps the most widely adopted pooling option in deep CNNs, 
which usually follows one or several convolutional layers to reduce the 
spatial dimensions of the feature maps. 
Let $\mathbf{x} \in R^{c\times h\times w}$ be the input feature map before a pooling
layer, where $c$ is the number of channels and $h$ and $w$ are the height
and width, respectively. A max pooling layer with pooling window of size 
$k\times k$ and stride $s \times s$ is defined by the function 
$\mathbf{z} = \mathcal{P}_k^s(\mathbf{x})$, where $\mathbf{z} \in R^{c \times \frac{h}{s} \times \frac{w}{s}}$, and
\begin{equation}
\label{eq:maxpool}
\begin{split}
&\mathbf{z}_{n, i, j} = \max_{\substack{i' \in [(i-1)s + 1, (i-1)s + k], i' \leq h \\ j'\in [(j-1)s + 1, (j-1)s + k], j'\leq w}} \mathbf{x}_{n, i', j'}, \\
&n \in [1, c], i \in [1, \frac{h}{s}], j \in [1, \frac{w}{s}].
\end{split}
\end{equation}
Specifically, to obtain the value at each spatial location of the output feature map $\mathbf{z}$, 
$\mathcal{P}_k^s(\cdot)$ selects the maximum activation within the corresponding local region 
of size $k\times k$ in $\mathbf{x}$. While performed in a single step, conceptually, max pooling can be  considered as two consecutive processes: 
\begin{equation}
\label{eq:maxpool_twostep}
\begin{split}
&\mathbf{o} = \mathcal{P}_k^1(\mathbf{x}), \; \; \mathbf{z} = \mathcal{D}^s(\mathbf{o}),
\end{split}
\end{equation}
where $\mathbf{z}_{n,i,j} = \mathbf{o}_{n, (i-1)s + 1, (j-1)s + 1}$.

In the first step, max pooling with window size $k\times k$ and stride $1\times 1$ is performed, 
producing an intermediate output $\mathbf{o}$, which has the same dimension as $\mathbf{x}$. 
In the second step, a spatial downsampling step is performed, where the value at the top left corner of each disjoint $s\times s$ window is selected to produce the output feature map 
with the spatial dimension reduced by $s$ times. The two-step view of max pooling allows us to 
investigate the differences of the effects of each step on learning. The first step $\mathcal{P}_k^1(\cdot)$ provides an additional level of nonlinearity to the CNN, as well as a certain degree of local (up to the scale of $k\times k$) distortion invariance. The second step $\mathcal{D}^s(\cdot)$, on the other hand, serves the purpose of reducing the amount of computation and weight parameters (given a fixed receptive field size) needed at upper layers of a deep CNN, as well as facilitating the model to learn more abstract representations by providing a more compact view of the input. 
We exploit this two-step view of the classical max pooling procedure and introduce a pooling algorithm which 
explicitly improves the downsampling step in order to learn models with better generalization ability.

\subsection{Pooling with Stochastic Spatial Sampling}
While the typical downsampling step of a max pooling layer intuitively reduces the spatial dimension of
a feature map by always selecting the activations at fixed locations, this design choice is somewhat 
arbitrary and potentially suboptimal. For example, as specified in Equation \ref{eq:maxpool_twostep}, 
the downsampling function $\mathcal{D}^s(\cdot)$ selects only the activation at the top left 
corner of each $s\times s$ disjoint window and discards the rest $s^2 - 1$ activations, which are equally 
informative for learning. Considering the total number of pooling layers present in a CNN, 
denoted by $L$, this deterministic downsampling approach discards $s^{2L} -1$ possible sampling choices. 
Therefore, although a natural design choice, deterministic uniform spatial sampling may not be optimal for the purpose of
learning where the goal is to generalize. On the other hand, if we allow the downsampling step to be performed in 
a non-uniform and non-deterministic way, where the sampled indices are not restricted to be at evenly 
distributed locations, we are able to produce many variations of downsampled feature maps. Motivated by this observation, 
we propose S3Pool, a variant of max pooling with a stochastic spatial downsampling procedure \footnote{Although we work with max pooling as the underlying pooling mechanism since it is widely used, the proposed S3Pool is oblivious to the nature of the first stage pooling and is applicable just as well to other types of pooling schemes (e.g., average pooling, stochastic pooling~\cite{zeiler2013stochastic}).}. S3Pool, denoted by
$\tilde{\mathcal{P}}^s_{k,g}(\cdot)$, works in a two-step fashion: the first step, $\mathcal{P}_k^1(\cdot)$,
is identical to max pooling, however, the second step, $\mathcal{D}^s(\cdot)$, is replaced by a stochastic version $\tilde{\mathcal{D}}^s_g(\cdot)$. 

Prior to the downsampling step of S3Pool, the feature map is divided into $\frac{h}{g}$ vertical and $\frac{w}{g}$ horizontal disjoint grids,
indexed by $p\in [1, \frac{h}{g}]$ and $q\in [1, \frac{w}{g}]$, respectively, with $g$ being the grid size. 
Within each vertical/horizontal grid, $\frac{g}{s}$ rows/columns are randomly chosen:
\begin{equation}
\begin{split}
\mathbf{r}^p = \mathrm{C}_{[(p-1)g+1, pg]}^{\frac{g}{s}}, \; \mathbf{c}^q = \mathrm{C}_{[(q-1)g+1, qg]}^{\frac{g}{s}},
\end{split}
\end{equation}
where $\mathrm{C}_{[a,b]}^{m}$ denotes a multinomial sampling function, which samples $m$ sorted integers 
randomly from the interval $[a, b]$ without replacement.
The indices drawn from each vertical/horizontal grid are then concatenated, producing a set of rows, $\mathbf{r}=[\mathbf{r}^1,\mathbf{r}^2,\cdots,\mathbf{r}^\frac{h}{g}]$
and a set of columns, $\mathbf{c}=[\mathbf{c}^1,\mathbf{c}^2,\cdots,\mathbf{c}^\frac{w}{g}]$, which leaves us the downsampled feature map being:
$\mathbf{z} = \tilde{\mathcal{D}}^s_g(\mathbf{o})$, where $\mathbf{z}_{n, i, j} = \mathbf{o}_{n, \mathbf{r}_i, \mathbf{c}_j}$. 
To summarize, given the grid size $g$, the stride $s$ and the pooling window size $k$, 
S3Pool is defined as:
\begin{equation}
\label{eq:s3pool}
\begin{split}
\mathbf{z} = \tilde{\mathcal{D}}^s_g(\mathcal{P}_k^1(\mathbf{x}))
\end{split}
\end{equation}

The grid size, $g$, is a hyperparameter of S3Pool which can control the level of stochasticity introduced. 
Figure~\ref{fig:diagram2} illustrates the effect of changing the grid size for 
the stochastic spatial downsampling $\mathcal{D}^2_g(\cdot)$. Larger grid sizes correspond to less uniformly sampled rows and columns. 
In the extreme case, where the grid size equals to the image size, S3Pool
selects $\frac{h}{s}$ rows and $\frac{w}{s}$ columns from the entire input feature map in a purely random fashion, 
which yields the maximum amount of randomness in sampling.

The behavior of $\tilde{\mathcal{D}}^s_g(\cdot)$ is intuitively visualized
using an image as input (Figure \ref{fig:cat}), which is downsampled by applying
uniform sampling, $\mathcal{D}^2(\cdot)$, and
stochastic downsampling with different grid sizes, $\tilde{\mathcal{D}}^2_{\frac{w}{4}}(\cdot)$, $\tilde{\mathcal{D}}^2_{\frac{w}{2}}(\cdot)$, $\tilde{\mathcal{D}}^2_{w}(\cdot)$. It can be seen that all the stochastic spatial sampling variants produce images that are recognizable to human eyes, with certain degrees of distortion, even in the extreme case where the grid size equals to the image size. The benefit of S3Pool is thus obvious in that, each draw from the pooling step will produce different yet plausible downsampled feature maps, which is equivalent to performing data augmentation \cite{simard2003best} at the pooling layer level. However, compared with traditional data augmentation, such as image cropping \cite{alexnet}, the distortion introduced by S3Pool is more aggressive. As a matter of fact, cropping
(which corresponds to horizontal and vertical translation) can be considered as a special case of S3Pool in the input layer, with $s=1$ and $g=w$, with the additional constraint that the sampled rows and columns are spatially contingent.

To further illustrate the idea of S3Pool and its difference from the standard max pooling, 
and another non-deterministic variant of max pooling~\cite{zeiler2013stochastic}, we demonstrate
the different pooling processes in Figure~\ref{fig:diagram1} using a toy feature map of size
$1\times 4 \times 4$. From the two-step view of max pooling, stochastic pooling~\cite{zeiler2013stochastic} modifies the first step: instead of outputing a deterministic maximum in each pooling window of $k\times k$, it randomly draws a response according to the magnitude of the activation; the second downsampling step, however, remains the same as in max pooling. Different from stochastic pooling~\cite{zeiler2013stochastic} and deterministic max pooling, S3Pool offers the flexibility to control the amount of distortion introduced in each sampling step by varying the grid size $g$ in each layer. This is useful especially for building deep CNNs with multiple pooling layers, 
which makes it possible to control the trade-off between the regularization strength and the converging speed.

In terms of implementation concerns, S3Pool does not introduce any additional parameters. 
It is easy to implement, and fast to compute during training time (in our experiments, we show that S3Pool introduces very little computational overhead compared to max pooling). 

{\bf Inference Stage.} During testing time, a straightforward but inefficient 
approach is to take the average classification outputs from 
many instances of CNN with S3Pool, which can otherwise act as a finite sample estimate of the expectation of S3Pool downsampling. A more efficient approach is to use the expectation of the downsampling procedure during testing.
The expected value at a location $(i,j)$ in the feature map (with $\widetilde{si} := ((i-1) \mod g/s) + 1, \widecheck{si}:= \lfloor s(i-1) / g\rfloor$, similarly for $sj$, and $i\in [h/s],\, j\in [w/s]$) is given as
\[
E[\mathbf{z}_{n,i,j}] = \sum_{a=\widetilde{si}}^{g-g/s+\widetilde{si}} ~~ \sum_{b=\widetilde{si}}^{g-g/s+ \widetilde{si}} w_{a,b} \mathbf{o}_{n, g \widecheck{si}+a, g \widecheck{sj}+b} ,
\]
where $w_{ab} = h_a h_b$ with $h_a=\binom{a-1}{\widetilde{si}-1}\binom{g-a}{g/s-\widetilde{si}}/\binom{g}{g/s}$ with the convention $\binom{0}{0}=1$ (similar for $h_b$ with $\widetilde{si}$ replaced with $\widetilde{sj}$). For $g=s$, this expectation reduces to average pooling over the $s\times s$ windows in the second downsampling step. For $g>s$, computing this expectation is expensive and cannot be easily parallelized in a GPU implementation, we thus still use average pooling with window and stride s in our experiments during testing as an approximation of this expectation. 
We also experimented with standard uniformly spaced downsampling at testing time (i.e., picking the top-left corner pixel), however this was consistently outperformed by average pooling, with negligible computational overhead. Hence, all the testing results of S3Pool in this paper are computed with average pooling over $s\times s$ windows.

\section{Experiments}
We evaluate S3Pool with three popular image classification benchmarks: CIFAR-10, CIFAR-100 and STL-10. Both CIFAR-10 and CIFAR-100 consist of $32\times 32$ color images, each with 50,000 images for training and 10,000 images for testing. STL-10 consists of $96\times 96$ colored images evenly distributed in 10 classes, with 5,000 images for training and 8,000 images for testing. All the three datasets have relatively few examples, which makes proper regularization extremely important. We note that it is not our goal to obtain state-of-the-art results on these datasets, but rather to provide a fair analysis of the effectiveness of S3Pool compared to other pooling and regularization methods.

\begin{table}[!tbh]
\small
\caption{The configurations of NIN and ResNet used on CIFAR-10 and CIFAR-100. Conv-c-d stands for a convolutional layer with $c$ filters of size $d\times d$. Pool-k-s stands for a pooling layer with pooling window $k\times k$ and stride $s\times s$. }
\label{tab:arch}
\begin{minipage}{0.5\linewidth}
\centering
\begin{tabular}{c}
NIN \\
\toprule
Conv-192-5 \\
Conv-160-1\\
Conv-96-1\\
\midrule
Pool-2-2\\
\midrule
Conv-192-5\\
Conv-192-1\\
Conv-192-1\\
\midrule
Pool-2-2 \\
\midrule
Conv-192-3\\
Conv-192-1\\
Conv-10-1 \\
\midrule

{Global Average Pooling} \\
\midrule
{Softmax} \\
\bottomrule
\end{tabular}
\end{minipage}%
\begin{minipage}{0.5\linewidth}
\centering
\begin{tabular}{c}
ResNet\\
\toprule
Conv-32-3\\
\midrule
$3 \times \left\{\begin{tabular}{@{\ }l@{}}
Conv-32-3 \\ Conv-32-3
\end{tabular}\right.$ \\
\midrule
{Pool-2-2} \\
\midrule
$3 \times \left\{\begin{tabular}{@{\ }l@{}}
Conv-64-3 \\ Conv-64-3
\end{tabular}\right.$ \\
\midrule
{Pool-2-2} \\
\midrule
$3 \times \left\{\begin{tabular}{@{\ }l@{}}
Conv-128-3 \\ Conv-128-3
\end{tabular}\right.$ \\
\midrule
Conv-10-1\\
\midrule
{Global Average Pooling} \\
\midrule
{Softmax} \\
\bottomrule
\end{tabular}
\end{minipage}
\end{table}

\begin{table*}[t]
\caption{Control experiments with NIN \cite{nin} on CIFAR-10 and CIFAR-100 (best seen in color).}
\label{tab:nin}
\centering
\small
\begin{tabular}{*7lc}
\toprule
\multirow{2}{*}{\textbf{Model}} & \multirow{2}{*}{\textbf{flip}}& \multirow{2}{*}{\textbf{crop}} & \multicolumn{2}{c}{CIFAR-10} & \multicolumn{2}{c}{CIFAR-100} & \multirow{2}{*}{sec/epoch}\\
\cline{4-7}
 & & & train err & test err & train err & test err &  \\
\midrule
NIN + dropout & N & N & 0.63 & 10.68 & 6.15 & 35.24 & \multirow{4}{*}{131}\\
\textcolor{red}{NIN + dropout} & \textcolor{red}{N} & \textcolor{red}{Y} & \textcolor{red}{1.62} & \textcolor{red}{10.11} & \textcolor{red}{11.64} & \textcolor{red}{34.08}\\
\textcolor{cyan}{NIN + dropout}  & \textcolor{cyan}{Y} & \textcolor{cyan}{N} & \textcolor{cyan}{1.28} & \textcolor{cyan}{9.75} & \textcolor{cyan}{8.57} & \textcolor{cyan}{33.48}\\
\textcolor{blue}{NIN + dropout} & \textcolor{blue}{Y} & \textcolor{blue}{Y} & \textcolor{blue}{2.67} & \textcolor{blue}{9.34} & \textcolor{blue}{14.15} & \textcolor{blue}{32.36}\\
\midrule
Zeiler et al.\cite{zeiler2013stochastic} & N & N & 0.01 & 12.86& 0.1 & 39.64 & \multirow{4}{*}{218} \\
\textcolor{red}{Zeiler et al.\cite{zeiler2013stochastic}} & \textcolor{red}{N} & \textcolor{red}{Y} & \textcolor{red}{0.06} & \textcolor{red}{10.97} &  \textcolor{red}{0.78} & \textcolor{red}{35.44}\\
\textcolor{cyan}{Zeiler et al.\cite{zeiler2013stochastic}} & \textcolor{cyan}{Y} & \textcolor{cyan}{N} & \textcolor{cyan}{0.02} & \textcolor{cyan}{10.47} &  \textcolor{cyan}{0.20} & \textcolor{cyan}{36.82}\\
\textcolor{blue}{Zeiler et al.\cite{zeiler2013stochastic}} & \textcolor{blue}{Y} & \textcolor{blue}{Y} & \textcolor{blue}{0.22} & \textcolor{blue}{9.14} &  \textcolor{blue}{1.54} & \textcolor{blue}{33.47}\\
\midrule
S3Pool-16-8 & N & N & 1.85 & 9.30  & 9.25 & 33.85 & \multirow{4}{*}{142}\\
\textcolor{red}{S3Pool-16-8} & \textcolor{red}{N} & \textcolor{red}{Y} & \textcolor{red}{2.86} & \textcolor{red}{8.77}  & \textcolor{red}{11.44} & \textcolor{red}{33.24}\\
\textcolor{cyan}{S3Pool-16-8} & \textcolor{cyan}{Y} & \textcolor{cyan}{N} & \textcolor{cyan}{3.26} & \textcolor{cyan}{8.04}   & \textcolor{cyan}{13.19} & \textcolor{cyan}{31.04}\\
\textcolor{blue}{S3Pool-16-8} & \textcolor{blue}{Y} & \textcolor{blue}{Y} & \textcolor{blue}{4.39} & \textbf{\textcolor{blue}{7.71}}  & \textcolor{blue}{16.66} & \textbf{\textcolor{blue}{30.90}}\\
\bottomrule
\end{tabular}
\end{table*}

\begin{table*}[t]
\caption{Control experiments with ResNet \cite{nin} on CIFAR-10 and CIFAR-100 (best seen in color).}
\label{tab:resnet}
\centering
\small
\begin{tabular}{*7lc}
\toprule
\multirow{2}{*}{\textbf{Model}} & \multirow{2}{*}{\textbf{flip}}& \multirow{2}{*}{\textbf{crop}} & \multicolumn{2}{c}{CIFAR-10} & \multicolumn{2}{c}{CIFAR-100} & \multirow{2}{*}{sec/epoch}\\
\cline{4-7}
 & & & train err & test err & train err & test err \\
\midrule
ResNet & N & N &0.00  &14.07  & 0.02 & 42.32 & \multirow{4}{*}{120}\\
\textcolor{red}{ResNet} & \textcolor{red}{N} & \textcolor{red}{Y} & \textcolor{red}{0.01} & \textcolor{red}{9.21} & \textcolor{red}{0.06} & \textcolor{red}{33.88}\\
\textcolor{cyan}{ResNet}  & \textcolor{cyan}{Y} & \textcolor{cyan}{N} & \textcolor{cyan}{0.00} & \textcolor{cyan}{11.14} & \textcolor{cyan}{0.02} & \textcolor{cyan}{36.05}\\
\textcolor{blue}{ResNet} & \textcolor{blue}{Y} & \textcolor{blue}{Y} & \textcolor{blue}{0.06} & \textcolor{blue}{7.72} & \textcolor{blue}{0.48} & \textcolor{blue}{30.88}\\
\midrule
Zeiler et al.\cite{zeiler2013stochastic} & N & N & 0.01 & 9.94 & 0.04 & 34.42 & \multirow{4}{*}{152}\\
\textcolor{red}{Zeiler et al.\cite{zeiler2013stochastic}} & \textcolor{red}{N} & \textcolor{red}{Y} & \textcolor{red}{0.04} & \textcolor{red}{8.60} & \textcolor{red}{0.27} & \textcolor{red}{33.16}\\
\textcolor{cyan}{Zeiler et al.\cite{zeiler2013stochastic}} & \textcolor{cyan}{Y} & \textcolor{cyan}{N} & \textcolor{cyan}{0.05} & \textcolor{cyan}{8.06} & \textcolor{cyan}{0.15} & \textcolor{cyan}{31.76}\\
\textcolor{blue}{Zeiler et al.\cite{zeiler2013stochastic}} & \textcolor{blue}{Y} & \textcolor{blue}{Y} & \textcolor{blue}{0.23} & \textcolor{blue}{8.58} & \textcolor{blue}{1.24} & \textcolor{blue}{30.09}\\
\midrule
S3Pool-16-8 & N & N & 0.82  &8.86  & 3.97 & 32.78 & \multirow{4}{*}{125}\\
\textcolor{red}{S3Pool-16-8} & \textcolor{red}{N} & \textcolor{red}{Y} & \textcolor{red}{1.47} & \textcolor{red}{8.48} & \textcolor{red}{7.24} & \textcolor{red}{32.21}\\
\textcolor{cyan}{S3Pool-16-8} & \textcolor{cyan}{Y} & \textcolor{cyan}{N} & \textcolor{cyan}{1.90} & \textcolor{cyan}{7.31} & \textcolor{cyan}{8.28} & \textcolor{cyan}{30.65}\\
\textcolor{blue}{S3Pool-16-8} & \textcolor{blue}{Y} & \textcolor{blue}{Y} & \textcolor{blue}{3.23} & \textbf{\textcolor{blue}{7.09}} & \textcolor{blue}{12.47} & \textbf{\textcolor{blue}{29.36}}\\
\bottomrule
\end{tabular}
\end{table*}

\subsection{CIFAR-10 and CIFAR-100} \label{sec:cifar}
For CIFAR-10 and CIFAR-100, we experiment with two state-of-the-art architectures, network in network (NIN) \cite{nin} and residual networks (ResNet) \cite{residualnet}, both of which are well established architectures, but with different designs. We apply identical architectures on CIFAR-10 and CIFAR-100, except for the top convolultional layer for softmax (10 versus 100). The architectures we use in this paper differ slightly from those in \cite{nin,residualnet}, which we summarize in Table \ref{tab:arch}. Here Conv-c-d denotes a convolutional layer with c filters of size $d \times d$; Pool-k-s denotes a pooling layer implementation with pooling window $k \times k$ and stride $s \times s$. Batch normalization \cite{batchnorm} is applied to each convolutional layer for each of the two models, with ReLU as the nonlinearity. 

For each of the two models, we experiment with three variants of the pooling layers: 

\textbf{Standard pooling}: for NIN, both of the two Pool-2-2 layers are max pooling with pooling window of size $2\times 2$ and stride $2 \times 2$; a dropout layer with rate $0.5$ is also inserted after each pooling layer. For ResNet, we follow the original design in \cite{residualnet} by replacing the Pool-2-2 layer with stride 2 convolution, without dropout.
 
\textbf{Stochastic pooling}: proposed by Zeiler et al. \cite{zeiler2013stochastic} with pooling window of size $2\times 2$ and stride $2 \times 2$.

\textbf{S3Pool}: the proposed pooling method with pooling window of size $2\times 2$ and stride $2 \times 2$. Grid size $g$ is set as $16$ and $8$ for the first and second S3Pool layer, respectively (that is, each feature map is divided into 2 vertical and horizontal strips). We denote this implementation of S3Pool as S3Pool-16-8.

In addition to experimenting with different network structures and pooling methods, we
also employ different data augmentation strategies: with or without horizontal flipping and without or without cropping \footnote{4 pixels are padded at each border of the $32\times 32$ images, and random $32 \times 32$ crops are selected at each forward pass.}. We train all the models with ADADELTA \cite{adadelta} with an initial learning rate of 1 and a batch size of 128. For all the NIN variants, training takes 200 epochs with the learning rate reduced to $0.1$ at the $150$-th epoch. All the ResNet variants are trained for a total of 120 epochs with the learning rate reduced to $0.1$ at the $80$-th epoch.  

The experimental results are summarized in Table~\ref{tab:nin} and Table~\ref{tab:resnet} for NIN and Resnet respectively. For each set of the experiments, we show the training and testing error of the final epoch (for S3Pool, an average pooling layer of pooling window and stride $2\times 2$ is added following each S3Pool layer). We also show the average training time of each pooling option when used with different networks, measured by the number of seconds per epoch (that is, the time taken for a full pass of the training data for weight updates, and a full pass of the testing data). 

We observe that for every combination of dataset type, network architecture and
data augmentation technique (denoted by rows with the same color in Table \ref{tab:nin} and Table~\ref{tab:resnet}), S3Pool achieves the lowest testing error, while yielding higher training errors than NIN with dropout, ResNet and their counterparts with stochastic
pooling~\cite{zeiler2013stochastic}. More remarkably, S3Pool without any data augmentation
can outperform other methods with data augmentation in most of cases. In particular, S3Pool without
data augmentation is able to outperform the baselines with cropping on all of the four dataset and architecture combinations. On CIFAR-10, S3Pool is even able to outperform image flipping and cropping augmented dropout version of NIN (9.30 versus 9.34). 
The high performance of S3Pool even without data augmentation is consistent with our
understanding of the stochastic spatial sampling step as an implicit data augmentation strategy. Interestingly, while both flipping and cropping are beneficial to S3Pool,
flipping seems to produce more performance gain than cropping. This is reasonable
since the stochastic downsampling step in S3Pool does not change the horizontal spatial 
order of sampled columns.

As for the computational cost, S3Pool increases the training time by $8\%$ and $4\%$ on NIN and ResNet, respectively. Stochastic pooling, on the other hand, yields a much higher computational overhead of $66\%$ and $27\%$, respectively \footnote{All models are implemented with Theano, and ran on a single NVIDIA K40 GPU.}. This demonstrates 
that S3Pool is indeed a practical as well as effective implementation choice when used in deep CNNs. 

\begin{figure*}[t]
\centering
\includegraphics[scale=0.8]{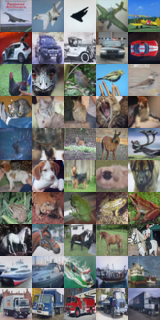}
\includegraphics[scale=0.8]{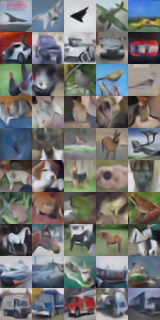}
\includegraphics[scale=0.8]{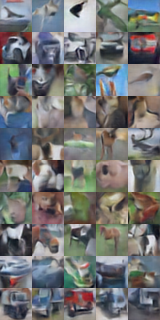}
\caption{Illustration of the behavior of S3Pool with deconvolutional neural networks on CIFAR-10 (best seen in color). From left to right: 50 images sampled from the test set, reconstructions obtained after the second pooling layer when using deterministic max pooling (center) and S3Pool (right). Note that even after two layers of stochastic spatial sampling, one is able to reconstruct recognizable images with various spatial distortions. }
\label{fig:deconv}
\end{figure*}

\begin{table}
\caption{Performance of different configurations of S3Pool by varying the grid sizes. All results are obtained with ResNet on CIFAR-10, without any data augmentation.}
\label{tab:gridsize}
\centering
\begin{tabular}{*3l}
\toprule
Configuration & train err & test err \\
\midrule
S3Pool-32-16 &2.58 & 9.32\\
S3Pool-16-8 &0.82 & \textbf{8.86}\\
S3Pool-8-8 & 1.29 & 10.14\\
S3Pool-8-4 & 0.92& 11.04\\
S3Pool-4-4 & 0.72& 11.02\\
S3Pool-2-2 & 0.26 &13.01 \\
\bottomrule
\end{tabular}
\end{table}

\paragraph{Effect of grid size}To investigate the effect of the grid size of S3Pool, we take the same ResNet architecture used in Section \ref{sec:cifar}, replace the S3Pool-16-8 layers
with different grid size settings, and report the results on CIFAR-10 in Table \ref{tab:gridsize}. We can observe that, in general, increasing the grid size of S3Pool yields larger training errors, as a result of more stochasticity; the testing error on the other hand, first decreases thanks to stronger regularization, then increases when the training error is too high. This observation suggests a trade-off between the optimization feasibility and the generalization ability, which can be adjusted in different applications by setting the grid sizes of each S3Pool layer.

\begin{figure}[!tbh]
\centering
\includegraphics[scale=0.4]{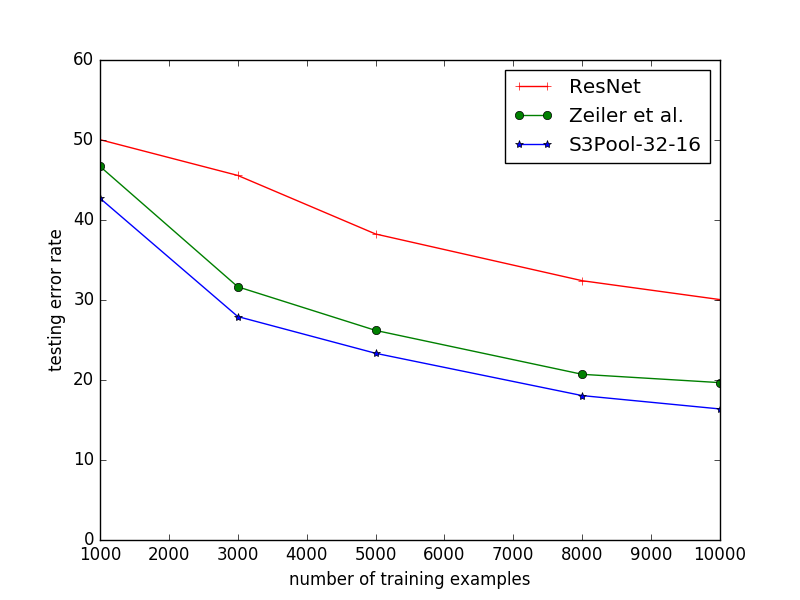}
\caption{Testing error rate on CIFAR-10 with different training data sizes (best seen in color).}
\label{fig:num_examples}
\end{figure}

\paragraph{Learning with limited training data}We further take the same ResNet architecture, and perform experiments with fewer training examples in CIFAR-10, which is shown in Figure \ref{fig:num_examples}. The results indicate that, by varying the number of training examples from as low as $1000$ to $10000$, S3Pool achieves consistently lower testing errors compared with the baseline ResNet as well as stochastic pooling~\cite{zeiler2013stochastic}.

\begin{table}
\caption{Results on STL-10. S3Pool-$g_1$-$g_2$-$g_3$-$g_4$ denotes the configuration
of the grid size at each of the four S3Pool layer. }
\label{tab:stl10}
\centering
\small
\begin{tabular}{*3lc}
\toprule
\textbf{model} & train err & test err & sec/epoch\\
\midrule
ResNet &0.00 &39.84 & 30\\
Zeiler et al. \cite{zeiler2013stochastic} &0.00 &25.93 & 70\\
\midrule
S3Pool-96-48-24-12 &2.12 & \textbf{24.06} & \multirow{6}{*}{35}\\
S3Pool-48-24-12-6 &1.04 & 25.36 &\\
S3Pool-24-12-6-4 &0.12 & 29.21 &\\
S3Pool-12-6-4-4 &0.12 & 30.01 &\\
S3Pool-4-4-4-4 &0.06 &29.60 &\\
S3Pool-2-2-2-2 &0.02 &35.14 &\\
\midrule
Zhao et al. \cite{zhao2015stacked} & - & 25.47 & \multirow{3}{*}{-}\\
Dosovitskiy et al. \cite{dosovitskiy2014discriminative} & - & 27.2 & \\
Yang et al. \cite{targetcoding} & - & 26.85 & \\
\bottomrule
\end{tabular}
\end{table}

\subsection{STL-10}
STL-10 has much fewer training examples 
and larger image sizes compared with CIFAR-10/CIFAR-100. We adopt the 18-layer ResNet based
architecture on this dataset, and test different pooling methods by replacing the stride 2
convolutions by stochastic pooling~\cite{zeiler2013stochastic} and S3Pool with different grid size
settings. 
We follow similar training protocols as in Section \ref{sec:cifar}, except that all the models are trained for 200 epochs with the learning rate decreased by a factor of $10$ at 
the $150$-th epoch, with no data augmentation applied.

The results are summarized in Table \ref{tab:stl10}. All variations of S3Pool significantly
improve the performance of the baseline ResNet. In particular,
S3Pool with the strongest regularization (S3Pool-96-48-24-12) 
achieves the state-of-the-art testing error on STL-10, outperforming supervised learning \cite{targetcoding} as well as semi-supervised learning \cite{zhao2015stacked,dosovitskiy2014discriminative} approaches. In terms of computational
cost, S3Pool only increases the training time by $16\%$ compared with the basic ResNet, even with four S3Pool layers.  

\subsection{Visualization}
Despite the convenient visualization of stochastic spatial sampling in the pixel space as shown in Figure \ref{fig:cat}, it is still unclear whether the same intuition holds when S3Pool is used in higher layers, and/or several S3Pool layers are stacked in a deep CNN. To this end, we obtain a trained NIN with two S3Pool layers as specified in Section~\ref{sec:cifar}, fix all the weights below the second S3Pool layer, turn off the stochasticity (i.e., using the test model of S3Pool) and stack a deconvolutional network \cite{deconv} on top. The output of the deconvolutional network is then trained to reconstruct the inputs from the training set of CIFAR-10 in a deterministic way. After training, we can sample reconstructions from the deconvolutional network with stochasticity. The results are shown in Figure \ref{fig:deconv}, where in the left column we show 50 images from the testing set, and each row shows the first 5 images from each of the 10 classes: airplane, automobile, bird, cat, deer, dog, frog, horse, ship, truck. The second column shows the reconstructions produced by the deconvolutional network with the test mode of S3Pool (no sampling). The third column shows the a single draw of the reconstructions from the network with S3Pool layers. Note that the third column gives different reconstructions at each run of the deconvolutional network, due to its stochastic nature.

It is noticed that by turning off the stochastic spatial sampling (second column), the deconvolutional network is able to faithfully reconstruct the shape and the location of the objects, subject to reduced image details. The reconstructions from the network with S3Pool are also visually meaningful, 
even with strong stochasticity (in this case, the grid sizes are set to $16$ and $8$ for the
two S3Pool layers). In particular, most reconstructions correspond to recognizable objects with various spatial distortions: local rescaling, translation, and etc.. Also note that these distortions do not follow a fixed pattern, thus can not be easily obtained by applying a basic geometric transform to the images directly. Therefore, the benefit of S3Pool can be understood as, during training, instead of using samples from the training set directly (first column in Figure \ref{fig:deconv}), the S3Pool layers sample locally distorted features (third column in Figure \ref{fig:deconv}) which are used implicitly for training. This corresponds to an aggressive data augmentation, which can significantly improve the generalization ability. The observation agrees with the results in Table \ref{tab:nin} and Table~\ref{tab:resnet}, where S3Pool outperforms all image cropping augmented baselines, as image cropping can be considered as a much milder data augmentation than S3Pool.

\section{Conclusions}
We proposed S3Pool, a novel pooling method for CNNs. S3Pool extends the standard max pooling by decomposing pooling into two steps: max pooling
with stride 1 and a non-deterministic spatial downsampling step by randomly sampling 
rows and columns from a feature map. 
In effect, S3Pool implicitly augments the training data at each pooling stage which enables 
superior generalization ability of the learned model. 
Extensive experiments on CIFAR-10 and CIFAR-100 have demonstrated that, S3Pool, either used in
conjunction with data augmentation or not, significantly outperforms standard max pooling, dropout, and an existing stochastic pooling approach. In particular, by adjusting the level of stochasticity introduced by S3Pool using a simple mechanism, we obtained state-of-art result on STL-10. Additionally, S3Pool is simple to implement and introduces little computational overhead compared to general max pooling, which makes it a desirable design choice for learning deep CNNs. 

{\small
\bibliographystyle{ieee}
\bibliography{egbib}

\begin{thebibliography}{10}\itemsep=-1pt

\bibitem{Coates11}
A.~Coates and A.~Y. Ng.
\newblock Selecting receptive fields in deep networks.
\newblock In {\em NIPS}, 2011.

\bibitem{Csurka04}
G.~Csurka, C.~Dance, L.~Fan, J.~Willamowski, and C.~Bray.
\newblock Visual categorization with bags of keypoints.
\newblock In {\em ECCV Workshop}, 2004.

\bibitem{dosovitskiy2014discriminative}
A.~Dosovitskiy, J.~T. Springenberg, M.~Riedmiller, and T.~Brox.
\newblock Discriminative unsupervised feature learning with convolutional
  neural networks.
\newblock In {\em NIPS}, 2014.

\bibitem{maxout}
I.~J. Goodfellow, D.~Warde-Farley, M.~Mirza, A.~Courville, and Y.~Bengio.
\newblock Maxout networks.
\newblock {\em ICML}, 2013.

\bibitem{fractional}
B.~Graham.
\newblock Fractional max-pooling.
\newblock {\em arXiv preprint arXiv:1412.6071}, 2014.

\bibitem{Kristen05}
K.~Grauman and T.~Darrell.
\newblock Pyramid match kernels: Discriminative classification with sets of
  image features.
\newblock In {\em ICCV}, 2005.

\bibitem{BengioPool14}
C.~Gulcehre, K.~Cho, R.~Pascanu, and Y.~Bengio.
\newblock Learned-norm pooling for deep feedforward and recurrent neural
  networks.
\newblock In {\em MLKDD}, 2014.

\bibitem{he2014spatial}
K.~He, X.~Zhang, S.~Ren, and J.~Sun.
\newblock Spatial pyramid pooling in deep convolutional networks for visual
  recognition.
\newblock In {\em ECCV}, 2014.

\bibitem{residualnet}
K.~He, X.~Zhang, S.~Ren, and J.~Sun.
\newblock Deep residual learning for image recognition.
\newblock {\em CVPR}, 2016.

\bibitem{john1996sampling}
J.~R. Higgins.
\newblock {\em Sampling theory in Fourier and signal analysis: foundations}.
\newblock Oxford University Press on Demand, 1996.

\bibitem{Hubel62}
D.~Hubel and T.~Wiesel.
\newblock Receptive fields, binocular interaction and functional architecture
  in the cats visual cortex.
\newblock {\em The Journal of Physiology}, 160:106--154, 1962.

\bibitem{batchnorm}
S.~Ioffe and C.~Szegedy.
\newblock Batch normalization: Accelerating deep network training by reducing
  internal covariate shift.
\newblock {\em ICML}, 2015.

\bibitem{alexnet}
A.~Krizhevsky, I.~Sutskever, and G.~E. Hinton.
\newblock Imagenet classification with deep convolutional neural networks.
\newblock In {\em NIPS}, 2012.

\bibitem{Lana06}
S.~Lazebnik, C.~Schmid, and J.~Ponce.
\newblock Beyond bags of features: Spatial pyramid matching for recognizing
  natural scene categories.
\newblock In {\em CVPR}, 2006.

\bibitem{Lecun98}
Y.~LeCun, L.~Bottou, Y.~Bengio, and P.~Haffner.
\newblock Gradient-based learning applied to document recognition.
\newblock {\em Proceedings of the IEEE}, 86(11):2278–--2324, 1998.

\bibitem{Lee16}
C.~Lee, P.~Gallagher, and Z.~Tu.
\newblock Generalizing pooling functions in convolutional neural networks:
  Mixed, gated, and tree.
\newblock In {\em AISTATS}, 2016.

\bibitem{nin}
M.~Lin, Q.~Chen, and S.~Yan.
\newblock Network in network.
\newblock {\em ICLR}, 2014.

\bibitem{Patch15}
X.~Lu, Z.~Lin, X.~Shen, R.~Mech, and J.~Wang.
\newblock Deep multi-patch aggregation network for image style, aesthetics, and
  quality estimation.
\newblock In {\em ICCV}, 2015.

\bibitem{shannon1949communication}
C.~E. Shannon.
\newblock Communication in the presence of noise.
\newblock {\em Proceedings of the IRE}, 1949.

\bibitem{simard2003best}
P.~Y. Simard, D.~Steinkraus, and J.~C. Platt.
\newblock Best practices for convolutional neural networks applied to visual
  document analysis.
\newblock In {\em ICDAR}, 2003.

\bibitem{pmaxout}
J.~T. Springenberg and M.~Riedmiller.
\newblock Improving deep neural networks with probabilistic maxout units.
\newblock {\em ICLR Workshop}, 2014.

\bibitem{dropout}
N.~Srivastava, G.~Hinton, A.~Krizhevsky, I.~Sutskever, and R.~Salakhutdinov.
\newblock Dropout: A simple way to prevent neural networks from overfitting.
\newblock {\em JMLR}, 2014.

\bibitem{TDP15}
G.~Xie, X.~Zhang, X.~Shu, S.~Yan, and C.~Liu.
\newblock Task-driven feature pooling for image classification.
\newblock In {\em ICCV}, 2015.

\bibitem{targetcoding}
S.~Yang, P.~Luo, C.~C. Loy, K.~W. Shum, and X.~Tang.
\newblock Deep representation learning with target coding.
\newblock In {\em AAAI}, 2015.

\bibitem{adadelta}
M.~D. Zeiler.
\newblock Adadelta: an adaptive learning rate method.
\newblock {\em arXiv preprint arXiv:1212.5701}, 2012.

\bibitem{zeiler2013stochastic}
M.~D. Zeiler and R.~Fergus.
\newblock Stochastic pooling for regularization of deep convolutional neural
  networks.
\newblock {\em ICLR}, 2013.

\bibitem{deconv}
M.~D. Zeiler, D.~Krishnan, G.~W. Taylor, and R.~Fergus.
\newblock Deconvolutional networks.
\newblock In {\em CVPR}, 2010.

\bibitem{zhao2015stacked}
J.~Zhao, M.~Mathieu, R.~Goroshin, and Y.~Lecun.
\newblock Stacked what-where auto-encoders.
\newblock {\em ICLR Workshop}, 2016.

\end{thebibliography}
}

\end{document}